\documentclass{ametsoc}

\journal{jhm}

\bibpunct{(}{)}{;}{a}{}{,}

\title{Effective Cloud Detection and Segmentation using a Gradient-Based Algorithm for Satellite Imagery; Application to improve PERSIANN-CCS}

    \authors{Negin Hayatbini\correspondingauthor{Negin Hayatbini, 
     Center for Hydrometeorology and Remote Sensing (CHRS), The Henry Samueli School of Engineering, Department
of Civil and Environmental Engineering, University of California, Irvine, Irvine, California.}, Kuo-lin Hsu, and Soroosh Sorooshian}

     \affiliation{Center for Hydrometeorology and Remote Sensing (CHRS), Department
of Civil and Environmental Engineering, University of California, Irvine, Irvine, California}

\email{nhayatbi@uci.edu}

    \extraauthor{Yunji Zhang and Fuqing Zhang}
    \extraaffil{Department of Meteorology, The Pennsylvania State University, University Park, Pennsylvania}

\abstract{Being able to effectively identify clouds and monitor their evolution is one important step toward more accurate quantitative precipitation estimation and forecast. In this study, a new gradient-based cloud-image segmentation technique is developed using tools from image processing techniques. This method integrates morphological image gradient magnitudes to separable cloud systems and patches boundaries. A varying scale-kernel is implemented to reduce the sensitivity of image segmentation to noise and capture objects with various finenesses of the edges in remote-sensing images. The proposed method is flexible and extendable from single- to multi-spectral imagery. Case studies were carried out to validate the algorithm by applying the proposed segmentation algorithm to synthetic radiances for channels of the Geostationary Operational Environmental Satellites (GOES-R) simulated by a high-resolution weather prediction model. The proposed method compares favorably with the existing cloud-patch-based segmentation technique implemented in the PERSIANN-CCS (Precipitation Estimation from Remotely Sensed Information using Artificial Neural Network - Cloud Classification System) rainfall retrieval algorithm. Evaluation of event-based images indicates that the proposed algorithm has potential to improve rain detection and estimation skills with an average of more than 45\% gain comparing to the segmentation technique used in PERSIANN-CCS and identifying cloud regions as objects with accuracy rates up to 98\%.} 

\begin{document}

\maketitle

%
\section{Introduction}
A more accurate representation of the clouds is crucial for a wide variety of applications. For instance, cloud coverage and type identification have been used for nowcasting to deliver accurate weather forecasts \citep{papin2002unsupervised}, rainfall and satellite precipitation estimates \citep{sorooshian2000evaluation,hong2004precipitation}, better assessment of climate models \citep[e.g.,][]{randall2003confronting, waliser2009cloud, liu2013reducing, li2016robust} and various other weather analysis and applications \citep{christodoulou2003multifeature, zhang2009cloud}.

Among all applications, reliable estimation of precipitation is critical to manage and predict water resources and climate studies \citep{aghakouchak2012near}. Satellite observations are a main source of global precipitation estimates due to their high spatio-temporal resolutions and coverage. One way to estimate precipitation is through using multiple wavelengths of geostationary (GEO) satellite including commonly used infrared (IR) and visible (VIS) wavelengths \citep{hsu1999estimation, nasrollahi2013artificial}. Despite the diurnal visible data, Infrared (IR) images collected by geostationary satellites can indicate the presence of clouds in the atmosphere during both daytime and nighttime with high spatiotemporal resolution and consecutive coverage \citep{kuligowski2002self}. Precipitation rate is then measured based on an indirect relationship between clouds albedo and cloud top temperature. Another popular source of satellite-based information is passive microwave (PMW) images from sensors onboard low-Earth-orbiting (LEO) satellites which is more directly related to the ground precipitation rate \citep{joyce2004cmorph,kidd2003satellite}.

An example for the concept of using LEO-PMW satellite data along with the GEO-IR-based data to provide global precipitation estimation at near real-time is the GPM (Global Precipitation Measurement) satellite. The NASA GPM program provides a key dataset called Integrated Multi-satellite Retrievals for GPM (IMERG). IMERG has been developed to provide half-hourly  global precipitation monitoring at 0.1x0.1 \citep{huffman2015nasa}. The satellite-based estimation of IMERG consists of three groups of algorithms including the Climate Prediction Center (CPC) morphing technique (CMORPH) from NOAA Climate Prediction Center (CPC) \citep{joyce2004cmorph}; The Tropical Rainfall Measuring Mission (TRMM) Multi-satellite Precipitation Analysis from NASA Goddard Flight Center (TMPA) \citep{huffman2007trmm}, and microwave-calibrated Precipitation Estimation from Remotely Sensed Information using Artificial Neural Network-Cloud Classification System (PERSIANN-CCS) \citep{hong2004precipitation}.

The purpose of this study is to improve the current in use version of PERSIANN-CCS in IMERG. PERSIANN-CCS consists of three general steps including, 1- segmentation of satellite cloud images into cloud patches, 2- feature extraction, and classification of cloud patches, 3- Rainfall mapping and estimation. This algorithm utilized IR-only data to indirectly estimate precipitation from the cloud top temperature and is associated with inherent uncertainties arising from different parts of the algorithm.

Despite all the recent efforts to improve precipitation estimation and rain no rain detection \citep{karbalaee2017bias,tao2016deep,nasrollahi2013artificial,mahrooghy2012enhanced,behrangi2009persiann} not much attention is paid to improve cloud extraction and identification part of PERSIANN-CCS algorithm. Considerable uncertainties result from the first and one of the most important steps of precipitation estimation from remotely sensed information which is cloud segmentation and identification. Almost all cloud segmentation algorithms employed by the IR-based rainfall estimation algorithms are based on the assumption that colder cloud top specify higher precipitation probabilities \citep{kidd2003satellite,huffman2007trmm, levizzani2001precipitation}. Nevertheless, cold cloud tops do not necessarily imply precipitation. Intense precipitation is correlated with colder clouds. However, the contrary relationship may not be true \citep{nasrollahi2013artificial}. In addition to this issue, orographically induced precipitation or precipitating warm clouds (e.g., stratiform) may cause precipitation, which is not easily identified with current algorithms \citep{joyce2004cmorph}.

Several algorithms have been developed to detect and segment clouds from satellite imageries, such as constant or changing threshold methods \citep{rossow1993cloud,stowe1999scientific,kriebel2003cloud, bendix2004cloud,rep200707,sun2016universal}, multidimensional histogram approaches \citep{karner2000multi}, neural networks \citep{yhann1995application,tian1999study, jang2006neural}, Haze Optimized Transformation (HOT) algorithm \citep{zhang2002image,zhang2003quantitative}, statistical and pattern recognition methods \citep{molnar1985retrieval,karner2000multi,murino2014cloud}, Markov Random Field formalism \citep{le2009use}, variational gradient-based fusion method \citep{li2012variational}, pixel-based seed identification and object-based region growing - the watershed segmentation \citep{sedano2011cloud,beucher1993segmentation, vincent1991watersheds}, major axis of a scatterplot and the invariant pixels detection \citep{lin2015radiometric}.

Among the cloud detection techniques, the threshold-based segmentation methods are the most broadly applied due to their simplicity, fast function and agreeable accuracy of cloud detection \citep{hagolle2010multi,jedlovec2008spatial,zhu2015improvement}. Threshold-base approaches had been used in many cloud detection algorithms including ISCCP (International Satellite Cloud Climatology Project) \citep{schiffer1983international}, APOLLO (AVHRR Processing scheme Over cLouds, Land and Ocean) \citep{gesell1989algorithm}, and CLAVR (CLouds from the Advanced Very High-Resolution Radiometer (AVHRR), NOAA) \citep{kriebel2003cloud}. The algorithm from ISCCP \citep{rossow1985isccp} considers only two conditions of cloudy and clear sky observed visible and infrared radiances. However, the missing clouds that is similar to clear conditions can be the source of uncertainty \citep{zhu2015improvement, sun2016universal}. The algorithm used in NOAA CLAVR \citep{stowe1991global} adopts a window of 2 by 2 pixels as the identification unit for the cloud detection. Based on the number of pixels that pass the test, the pixel matrix is categorized in different classes of cloud, mixed, or clear sky. In this algorithm, the bright background can cause uncertainty \citep{greenhough2005towards,kriebel2003cloud,lin2015radiometric,sun2016universal}. The APOLLO algorithm \citep{saunders1988improved,gesell1989algorithm} identifies cloudy pixels using the five full-resolution AVHRR channels based on a temperature threshold in each channel. Based on the data, a pixel is recognized as a cloud pixel when the reflectance is higher than a definite threshold or has a lower value than the threshold set by temperature \citep{sun2016universal}.

In threshold-based techniques, clouds are in general differentiated by a higher reflectance and lower temperature than the background or earth surface. A major source of error comes from the complex land surface composition and high variability of reflectivity in different cloud types; a threshold that is appropriate to a certain cloud type or a certain geographical region may not be applicable for another \citep{sun2016universal}. Moreover, pixels defined by the threshold allow us to only consider the radiometric and textural features of an individual pixel rather than contextual information provided by the image regions as objects \citep{blaschke2014geographic}. Hence, the potential benefit of an automated object-based approach designed to segment the images into meaningful “objects” which can be described as a set of features and is independent of predetermined threshold is indispensable.

The other challenge with segmentation methods of precipitation retrieval algorithms, despite extensive developments in this area, is that they are defined for intensity-based single spectral images, and are rarely used for multispectral imageries \citep{soille2002advances}. Infrared (IR) images collected by geostationary satellites are able to indicate the presence of clouds in the atmosphere during both daytime and nighttime and are useful for the continuous monitoring of cloud properties \citep{xu2005statistical}. However, nearly all IR-based rainfall estimation algorithms suffer from a major source of error that comes from the sometimes false assumption that the colder the cloud top temperature, higher the probability of rain production. Additional measurements from different sensors could be taken into account to provide information about the structure and vertical profile of clouds that lead to rain \citep{xu2005statistical}. 

Furthermore, a technique for cloud identification based on utilization of multi-channel brightness temperature is expected to outperform single-channel information \citep{mecikalski2006forecasting,li2010improving}. For instance, Behrangi et al. (2009b) used the multispectral information to improve Rain/No Rain detection capabilities using a neural network-based framework. They found that using the combination of any two IR channels seems superior to the use of any single IR channel. Also, Gonzalez et al. (2012) segmented multispectral images from MSG–SEVIRI (Meteosat Second Generation–Spinning Enhanced Visible and Infrared Imager) using an order-invariant watershed algorithm applied to the gradient images. Once image objects have been segmented, they are classified using a multi-threshold method. A drawback for this method is that a fixed threshold on each reflectance images and brightness temperatures is needed for the classification of clouds \citep{gonzalez2012watershed}. In summary, the limitations of previous studies emphasize the need for a hybrid data-driven algorithm for cloud detection based on the multi-channel information. Focusing on the aforementioned issues that make cloud detection in satellite observations a challenge, this paper introduces an automated data-driven cloud detection algorithm framework based on the image gradient of the hyperspectral and high-spatial-resolution remote sensing data.

This work is motivated by the recognition of the potential ability of object-based segmentation methods to identify warm clouds with higher cloud-top brightness temperatures, which sometimes are difficult to handle by existing rainfall estimation algorithms (e.g., PERSIAN-CCS). In the experiments, the algorithm is applied to the model simulated GOES-16 ABI imagery, along with the simulated radiance for different channels of the GOES-16 ABI as the reference. 

The proposed algorithm is a mathematical morphology-based method. This algorithm is comprised of several approaches that have been optimized to extract the information from geostationary satellite imagery. It is capable of being extended to other types of data as well. By integrating the complementary information from multi-channel measurements into the segmentation algorithm, it is possible to detect clouds more efficiently and accurately. The proposed algorithm makes it possible to segment additional information about more varied cloud types that can then be fed into a precipitation estimation algorithm. Not all clouds will rain, but having information about specific cloud regions allows one to apply different assumptions or take environmental factors into account about the probability of rain for those regions.

\section{Methodology}
The Gradient-based Multi-Spectral Segmentation (GMS) algorithm mainly carries out the following functions: multi-scale gradient computation, markers generation, and segmentation.

\subsection*{GMS Cloud detection algorithm}
\begin{figure}[h]
\centerline{\includegraphics[width=19pc]{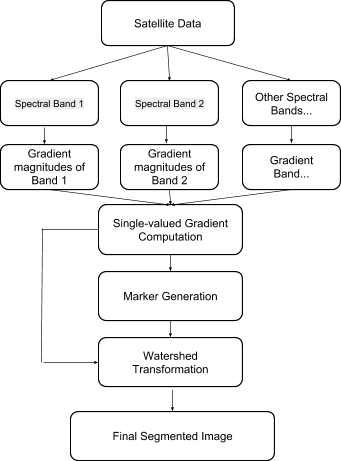}}
 \caption{Flow diagram of the proposed multi-spectral gradient-based segmentation algorithm.}\label{f1}
\end{figure}

The presented cloud detection approach is built on a hierarchical structure, consisting of different steps which the flow diagram of the proposed segmentation algorithm is shown in Fig. \ref{f1}.

\subsubsection*{Multiscale gradient magnitude computation}
Gradient computation here means achieving the maximum variation of each pixel's intensity in the neighborhood. The gradient highlights the sharp changes in intensity or the edges in an image. In a grayscale morphology, the gradient can be attained by subtracting the eroded image from the dilated image using a structuring element \citep{soille2002advances}. A gray-scale image pixels' value can be represented by the x and y coordinates as a three-dimensional set \citep{parvati2008image}. With this concept, gray-scale dilation can be defined as follows:

Let f(s, t) represent an image and B(x, y) be the structuring element. Structuring element or kernel is a group of pixels of different sizes and shapes that in this study a flat kernel which means a squared window of pixels with equal values are considered for simplicity. Gray-scale dilation of f by B is defined as below:
\begin{equation} \label{1}
\begin{aligned}
  &\mathrm(f\oplus B)_{(s, t)} = Max \Big\{f(s-x, t-y)+B(x, y)\mid(s-x), (t-y)\in D_f;(x, y)\in D_B\Big\} 
\end{aligned}
\end{equation}
Gray-scale erosion of f by B is defined as below:
\begin{equation} \label{2}
\begin{aligned}
  &\mathrm(f\ominus B)_{(s, t)} =
   &Min \Big\{f(s+x, t+y)-B(x, y)\mid(s+x), (t+y)\in D_f;(x, y)\in D_B\Big\} 
\end{aligned}
\end{equation}

where $D_f$ and $D_B$ are the domains of f and b, respectively. The conditions that (s+x) and (t+y)
have to be in the domain of f, and (x, y) have to be
in the domain of B, is equivalent to the condition in the
binary description of dilation and erosion. In dilation, two sets have to at least share one pixel in common and in erosion, structuring element
has to be completely contained by the set being eroded \citep{pahsa2006morphological}. For further details on the erosion and dilation operation please refer to the 'Digital Image Processing' book \citep{gonzalez1992digital}.

Gradient image is calculated using \eqref{3} from the original image and corresponds to the sharpness of intensity change for each pixel \citep{parvati2008image}. 
\begin{equation} \label{3}
\begin{aligned}
  &\mathrm{MG}(f) = (f\oplus B)-(f\ominus B)
\end{aligned}
\end{equation}
A multi-scale gradient algorithm capable of utilizing a varying scale-structuring element in mathematical morphology is implemented to reduce the sensitivity to noise and to extract various finenesses of the edges of the objects in remote sensing images \citep{wang1997multiscale}.
\begin{equation} \label{4}
\begin{aligned}
  &\mathrm{MG}(f) = \frac{1}{n} \sum_{i=1}^{n} [((f\oplus B_{i})-(f\ominus B_{i}))\ominus B_{i-1}]
\end{aligned}
\end{equation}
Where $B_i$ denotes the group of square structuring elements with the size (2i+1)$\times$(2i+1) pixels and n is the scale which in this study is set to 5. The mathematical morphology can be extended from a grayscale image to a multichannel image. By calculating the gradient magnitudes for each band of the multispectral image separately and combining them, the complementary measurements from other spectral bands are also considered. A component-wise strategy is utilized in this study to combine the resulting gradient image values. This approach consists of processing each channel of the multispectral image separately and summing up the resulting gradient image values to generate a single band gradient image.
\begin{equation} \label{5}
\begin{aligned}
  &\mathrm{MG}(f) = \sum_{i=1}^{n}{MG}(f_{i})
\end{aligned}
\end{equation}

\subsubsection*{Watershed segmentation on the gradient image}
\begin{figure}[h]
\centerline{\includegraphics[width=19pc]{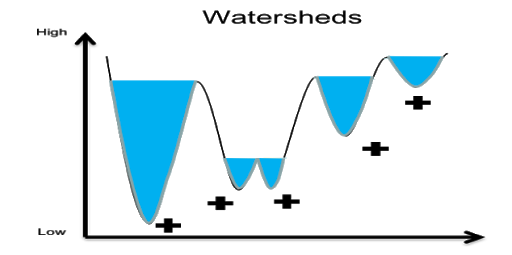}}
 \caption{Watershed Transformation Process Diagram. Each of the positive signs represents one watershed domain.}\label{f2}
\end{figure}

Watershed transformation \citep{vincent1991watersheds} is a powerful segmentation algorithm from mathematical morphology which have been used in many segmentation problems \citep{hsu2010extreme,zahraei2013short,lakshmanan2009efficient}. This approach is usually applied on Satellite images to extract the regions (i.e., objects) that are identified as clouds. The basic idea of the watershed algorithm is to consider the single channel intensity image as a three-dimensional topography map where the catchment basins are delimited by watershed lines. Each local minimum then flooded to neighboring pixels until meeting an adjacent catchment. A ridgeline is then delineated between any two regions’ borders (Fig. \ref{f2}). Further information about the watershed segmentation algorithm can be found in \citep{vincent1991watersheds}.

\begin{figure}[h]
\centerline{\includegraphics[width=39pc]{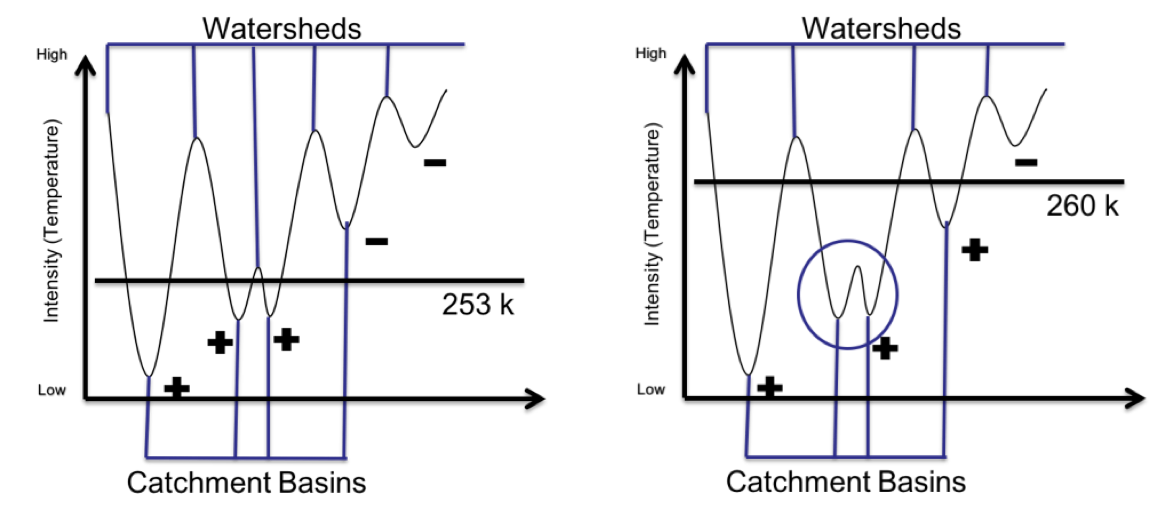}}
 \caption{Effect of increasing the threshold on the Watershed segmentation technique. The positive (negative) signs are representing the seeds that are included (excluded) in the segmentation process.}\label{f3}
\end{figure}

The watershed algorithm is a powerful morphological edge extractor and capable of separating overlapping clouds \citep{hong2006satellite}. The morphology-based watershed transformation used in this study is applied to the gradient magnitude of satellite images. This is as opposed to the traditional segmentations that apply watershed transformation directly on the imageries’ values. The inspiration is to capture warm clouds associated with rainfall in precipitation retrieval algorithms that are missed by utilization of traditional temperature threshold and watershed transformation methods. Fig. \ref{f3} is demonstrating how local minima associated with warmer cloud regions will be disregarded by the threshold applied to the values before implementing the watershed algorithm. In order to capture the clouds with the higher temperature that are associated with rainfall, the threshold-based algorithms use incremental series of thresholds to encompass a wider range of cloud types. However, as shown in schematic Fig. \ref{f3}, by increasing the temperature threshold from 253 to 260 for example, local minimas which here are representative of distinct clouds will merge and cause misleading outcomes. Therefore, optimal results are typically achieved when the watershed algorithm is applied to the gradient magnitude images that delineate the boundaries of all types of clouds, regardless of their height or temperature. Applying the watershed segmentation to gradient images may lead to over-segmentation due to the noise and slight changes in the local gradient calculations. Hence, a proper marker-generation technique is needed to achieve a desirable segmentation.

\subsubsection*{Marker generation and local minima elimination}
\begin{figure}[h]
\centerline{\includegraphics[width=19pc]{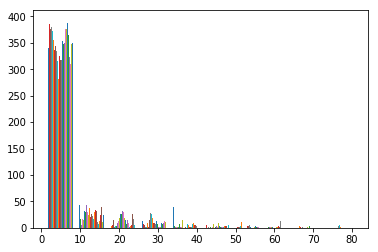}}
 \caption{Histogram plot of the gradient magnitude image. Horizontal axis is showing the range of gradient magnitude values, and the vertical axis is the number of pixels with each gradient magnitude value.}\label{f4}
\end{figure}

As described in the previous section, the high sensitivity of the watershed transformation algorithm to noise and irrelevant local minima yields many catchment basins, leading to over-segmentation. In order to overcome the over-segmentation issue only selected regional minima of the input image, called as “markers”, are being used prior to the application of the watershed transform. The markers are considered as either single points or larger regions that are placed inside an object of interest \citep{parvati2008image}. As a result, the number of regions in the watershed algorithm is reduced greatly. 
The generation of pertinent markers is one key step in the successful application of watershed segmentation \citep{soille2002advances}. In this study, we obtained markers through thresholding gradient magnitudes. Clusters of image pixels are classified as seed or non-seed pixels using Otsu thresholding method \citep{otsu1979threshold} to generate marker image. Otsu thresholding method is an optimal threshold which is selected automatically, based on maximizing the separability of the group-wise pixel values in gray levels. This approach selects an adequate threshold level for extracting objects from their background based on a gray-level histogram. The reason to implement this method is based on the histogram plot of the gradient magnitudes. As shown in figure \ref{f4}, gradient magnitudes of most pixels distribute in the lower value range while the sharp edge pixels with large gradient magnitudes are greater and separable than those within background. For further explanation of this thresholding method and examples, one can refer to \citep{otsu1979threshold,zhang2014marker}.

\subsection*{PERSIANN-CCS' segmentation algorithm}
PERSIANN-CCS is an infrared (IR) based algorithm being integrated into the IMERG (Integrated Multi-Satellite Retrievals for the Global Precipitation Mission GPM) to create a precipitation product in 0.1x0.1 degree resolution over the chosen domain 50N to 50S with 30 minutes time intervals. Although PERSIANN-CCS has a high spatial and temporal resolution, it may sometimes overestimate or underestimate due to the limitations associated with different parts of it \citep{karbalaee2017bias}. PERSIANN-CCS consists of three general steps including, 1- segmentation of satellite cloud images into cloud patches, 2- feature extraction, and classification of cloud patches, 3- Rainfall mapping and estimation. 

Image segmentation is the first, and most important step in all precipitation retrieval algorithms. Infrared-based rainfall estimation algorithms are categorized into the three general groups depending on the method used for extracting information from infrared cloud images: (a) pixel-based, (b) local texture–based, and (c) patch-based algorithms \citep{hong2004precipitation}. PERSIANN-CCS uses cloud-patch-based technique and applies artificial neural networks to classify clouds based on the IR information to estimate precipitation \citep{hong2004precipitation}. This algorithm outline clouds from the clear sky using a user-defined temperature threshold and segments cloud patches from geostationary IR images using a region-growing method \citep{gonzalez2007image,hong2006satellite}.

In the PERSIANN-CCS segmentation procedure, the minimum brightness temperature (Tbmin) of the clouds is first determined and then used as seed. Next, the temperature threshold is raised to identify a new set of pixels. Then, Tbmin is increased to include the neighboring areas from the seeded points. This procedure continues until reaching the borders of other seeded regions or cloud-free regions. The temperature threshold is iteratively increased to a maximum of 253 K. Afterward, a morphological operation is applied to remove/merge the tiny regions \citep{gonzalez2007image,hong2004precipitation}. Further detailed description of the algorithm along with the segmentation process can be obtained from \citep{hong2006satellite,hong2003precipitation}.

The rainfall estimation process in this algorithm is based on the extracted information from IR channels at three different temperature threshold levels (220, 235, and 253k) \citep{hong2006satellite}. PERSIANN-CCS is exclusively based on infrared data to identify the clouds and indirectly estimate rainfall from this single channel which can cause missing detection of the rainfall from warm clouds and false estimation for no precipitating cold clouds \citep{karbalaee2017bias}. In this study, the segmentation approach used in PERSIANN-CCS algorithm is applied to the simulated synthetic cloud imageries and the accuracy of cloud detection is compared with the proposed gradient-based segmentation algorithm.

\section*{Description of the datasets}

\subsection*{GOES-16 ABI}
The primary data sets collected and processed in this research include different channels of newest-generation geostationary weather satellites bands of GOES-R (Geostationary Operational Environmental Satellite, R series;). GOES-16 is the next generation of the GOES with the Advanced Baseline Imager (ABI; \citep{schmit2005introducing}) that has 16-channel imager including 2 visible channels, 4 near-infrared channels, and 10 infrared (IR) channels. Each of the channels is sensitive to a certain part of the land and the atmosphere and will give a better insight on the structure, type, and properties of the cloud. GOES-16 ABI provides more accurate, detailed, and timely detection of high-impact environmental phenomena by more spectral channels, higher spatial (2 km $\times$ 2 km at nadir for infrared channels) and temporal (every 5 minutes for CONUS) resolutions over the previous-generation operational GOES imager. The development of the ABI is being done as a collaborative effort between the National Aeronautics and Space Administration (NASA) and NOAA. For readers interested in further details, \citep{schmit2005introducing} and \citep{schmit2005introducing} are considered as the references. 


\subsection*{PSU WRF-EnKF dataset}

The Weather Research and Forecasting (WRF) model with its Advanced Research WRF (ARW) dynamical core, version 3.8.1 \citep{skamarock2008description} is used to generate the simulation for the synthetic infrared cloud imageries that are used to test and verify the proposed segmentation algorithm. This model is a community NWP model that is widely used in meteorological research and operational works. The applied model domain consisting 401 $\times$ 301 $\times$ 61 grid points with a horizontal resolution of 1 km; among the 61 vertical layers, there are 19 layers in the lowest 1 km above ground level, and the uppermost layer is located at 50 hPa. In order to deal with sub-grid physical processes that model grids are unable to resolve several physical parameterization schemes including the \citep{thompson2008explicit}, unified Noah land surface model \citep{ek2003implementation}, Monin-Obukhov Janjic Eta scheme \citep{chen1997impact} for surface layer parameterization, Mellor-Yamada-Janjic TKE scheme \citep{janjic1994step} for PBL processes, and the Rapid Radiative Transfer Model for General Circulation Models (RRTMG) schemes \citep{iacono2008radiative} for longwave and shortwave radiation are applied based on the sensitivity tests.

The initial conditions for the simulation are generated using the WRF-based ensemble Kalman filter (WRF-EnKF) data assimilation system developed at the Pennsylvania State University (PSU) \citep{zhang2009cloud,zhang2011performance}. This model-assimilated channel-10 brightness temperature observations from GOES-16 ABI for 1 hour from 1900 UTC to 2000 UTC every 5 minutes \citep{zhang2018assimilating}, which is a real-time 3-km atmospheric model covering CONUS and operated by the Earth System Research Laboratory (ESRL) of NOAA. The simulation is carried out for 4 hours and ends at 0000 UTC of the 13th of June 2017 with output file generated every 5 minutes. Simulated ABI brightness temperature fields are then generated from model output files using the Community Radiative Transfer Model (CRTM; \citep{han2006jcsda}). For further information on the PSU WRF-EnKF model simulation, and the synthetic cloud imageries along with the companion simulated radiance for different channels of the GOES-16 ABI please refer to \citep{zhang2018assimilating}.

\section*{Results and Discussion}
Experiments are carried out to validate the GMS algorithm by applying it to the model simulation of the GOES-16 ABI imageries, and the accompanying model-simulated horizontal distribution of all clouds hydrometeors in the model as the known ‘ground truth’. 
The first case study is a one-day period starting from 00:00 UTC of the 26th to 00:00 UTC of the 27th of August 2017 right after Hurricane Harvey made landfall in South Texas which brought record-breaking catastrophic rainfall in this region. Harvey is the second costliest hurricane (after Katrina in 2005) in insured losses; it is the first storm that was fully captured by the newest-generation geostationary satellites. The GOES-16 ABI radiances were recently successfully assimilated into the PSU WRF-EnKF data assimilation system with 3-km horizontal grid spacing while led to much improved subsequent WRF forecast of Harvey’s track and intensity \citep{minamide2018assimilation}. This 3-km WRF forecast output is used here to generate synthetic GOES-16 ABI radiances to provide case of validation of the proposed segmentation technique. The other case selected is a realistic simulation of the severe weather event across Wyoming and Nebraska on the 12th of June 2017 by a high-resolution state-of-the-science numerical weather prediction (NWP) model. It represents deep convective clouds over southeast Wyoming and the western Nebraska Panhandle, starting from 20:05 UTC of the 12th of June 2017 to 00:00 UTC of the 13th for every 5 minutes time interval. This is the first known tornadic thunderstorm outbreak that was well captured by the GOES-16 ABI observations and have been successfully assimilated into the PSU WRF-EnKF system as a convection-allowing model \citep{zhang2018assimilating}. 

Visual comparison and statistical evaluation are performed for both cases to obtain the accuracy of the proposed segmentation algorithm in comparison to the Single-channel, Threshold-based segmentation approach that is currently in use for precipitation retrieval algorithm, PERSIANN-CCS. Although the proposed gradient-based segmentation algorithm is capable of integrating information from different spectral bands, the results shown in this section are obtained from single-band IR cloud top brightness temperature to keep the consistency with the PERSIANN-CCS single-band segmentation algorithm. 

\subsection*{Visual comparison}

\begin{figure}[h]
\centerline{\includegraphics[width=39pc]{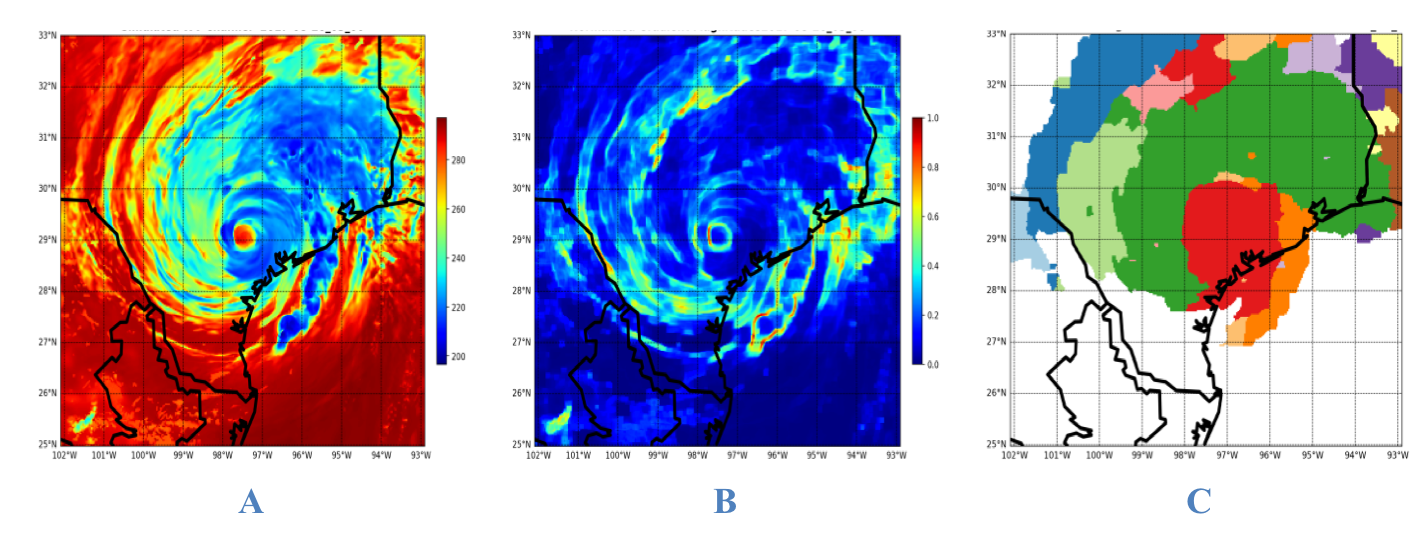}}
 \caption{Visualization of the gradient based segmentation result for simulated Hurricane Harvey event on August 26th 2017 at 3:00 AM UTC: A. Simulate IR longwave window band from GOES-16 ABI (The scale bar is temperature in degree Kelvin) B. Normalized gradient magnitudes imagery for the corresponding Simulated IR Channel (Scale bar is normalized gradient magnitude values). C. Resulting cloud Segments (Each random color represents a distinct segment).}\label{f6}
\end{figure}

\begin{figure}[h]
\centerline{\includegraphics[width=39pc]{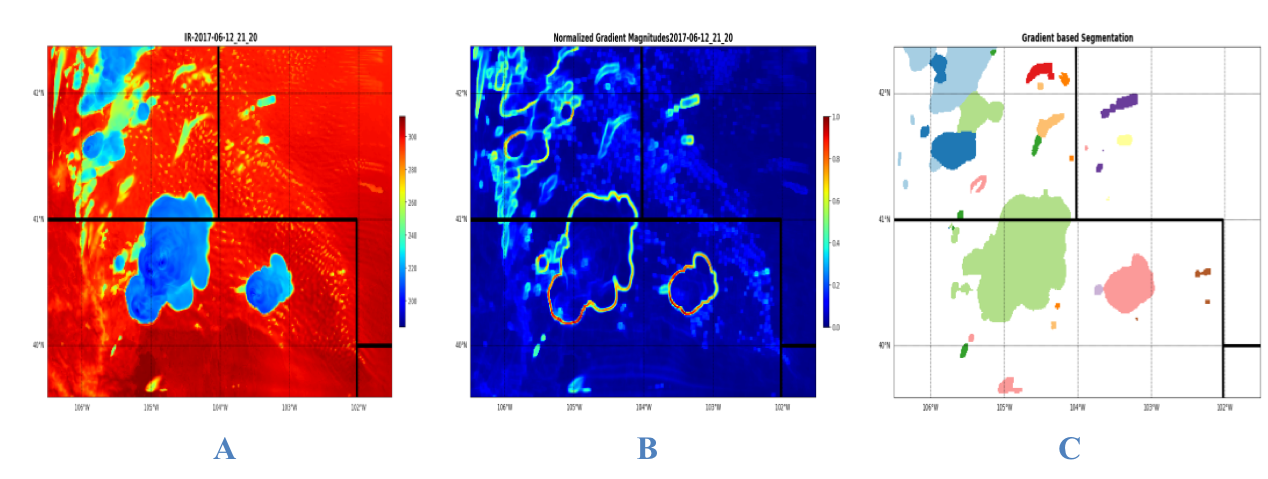}}
 \caption{ Visualization of the gradient based segmentation result for the simulated Wyoming tornado event at 21:20 UTC of the 12th of June 2017: A. Simulate IR longwave window band from GOES-16 ABI (The scale bar is temperature in degree Kelvin) B. Normalized gradient magnitudes imagery for the corresponding Simulated IR Channel (Scale bar is normalized gradient magnitude values). C. Resulting cloud Segments (Each random color represents a distinct segment).}\label{f8}
\end{figure}

The final gradient-based segmentation results from simulated IR input along with the gradient magnitude imageries of both the Hurricane Harvey, and Wyoming thunderstorm event are shown in Fig \ref{f6}, and \ref{f8} respectively. For these figures, the gradient magnitude is calculated from the IR image and the watershed segmentation is then applied to the gradient magnitude based on the generated markers to achieve the final cloud patches segmentation.

\begin{figure}[h]
\centerline{\includegraphics[width=39pc]{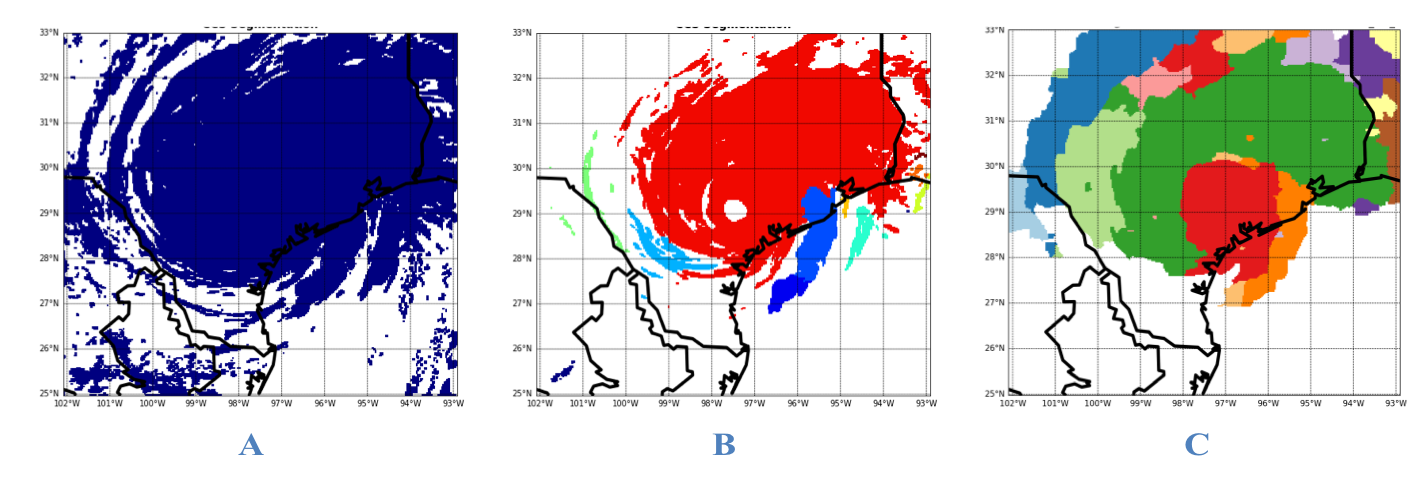}}
 \caption{Visual comparison of the two segmentation output (B and C) based on the 'truth' mask as a reference for the simulated Hurricane Harvey event at 3:00 AM UTC of August 26th 2017: A. 'Truth' Cloud Mask used as a reference. Dark blue region is implying the cloud existence. B. PERSIANN-CCS segmentation result from single-IR channel. C. Gradient-based Segmentation Algorithm output based on only IR channel. In panels B and C each random color identifies a distinct cloud patch.}\label{f7}
\end{figure}

\begin{figure}[h]
\centerline{\includegraphics[width=39pc]{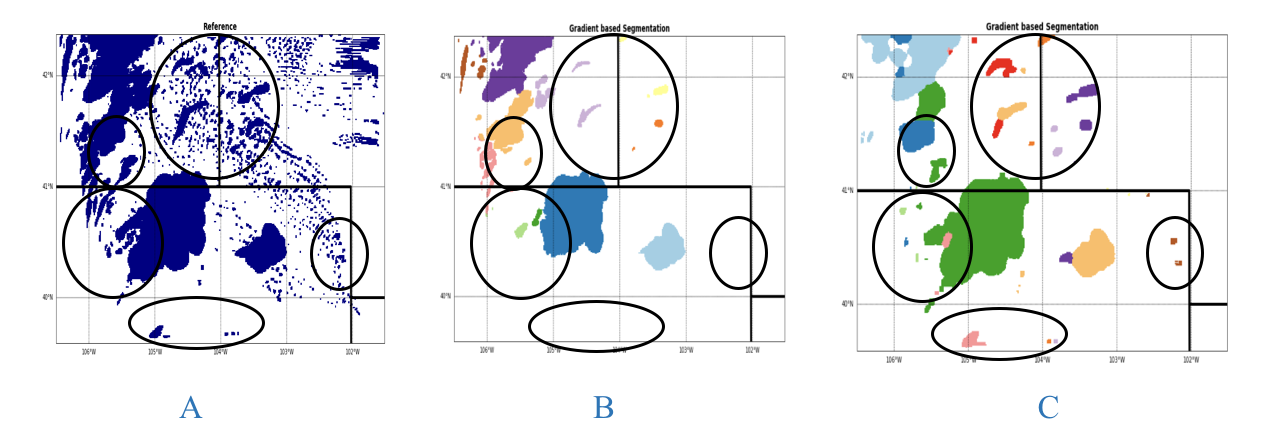}}
 \caption{Visual comparison of the two segmentation output (B and C) based on the 'truth' mask as a reference for the the simulated Wyoming tornado event at 21:20 UTC of the 12th of June 2017: A. 'Truth' Cloud Mask used as a reference. Dark blue region is implying the cloud existence. B. PERSIANN-CCS segmentation result from single-IR channel. C. Gradient-based Segmentation Algorithm output based on only IR channel. In panels B and C each random color identifies a distinct cloud patch.}\label{f9}
\end{figure}

Considering the reference images, in both cases the newly developed algorithm is capturing more of the clouds especially the warmer ones in comparison to the PERSIANN-CCS segmentation approach (Fig. \ref{f7}, and \ref{f9}). This indicates that the gradient-based segmentation algorithm is capable of overcoming the drawback associated with threshold-based segmentation approaches implemented in patch-based precipitation retrieval algorithms by capturing warmer clouds as well as the cold convective ones.

\begin{figure}[h]
\centerline{\includegraphics[width=39pc]{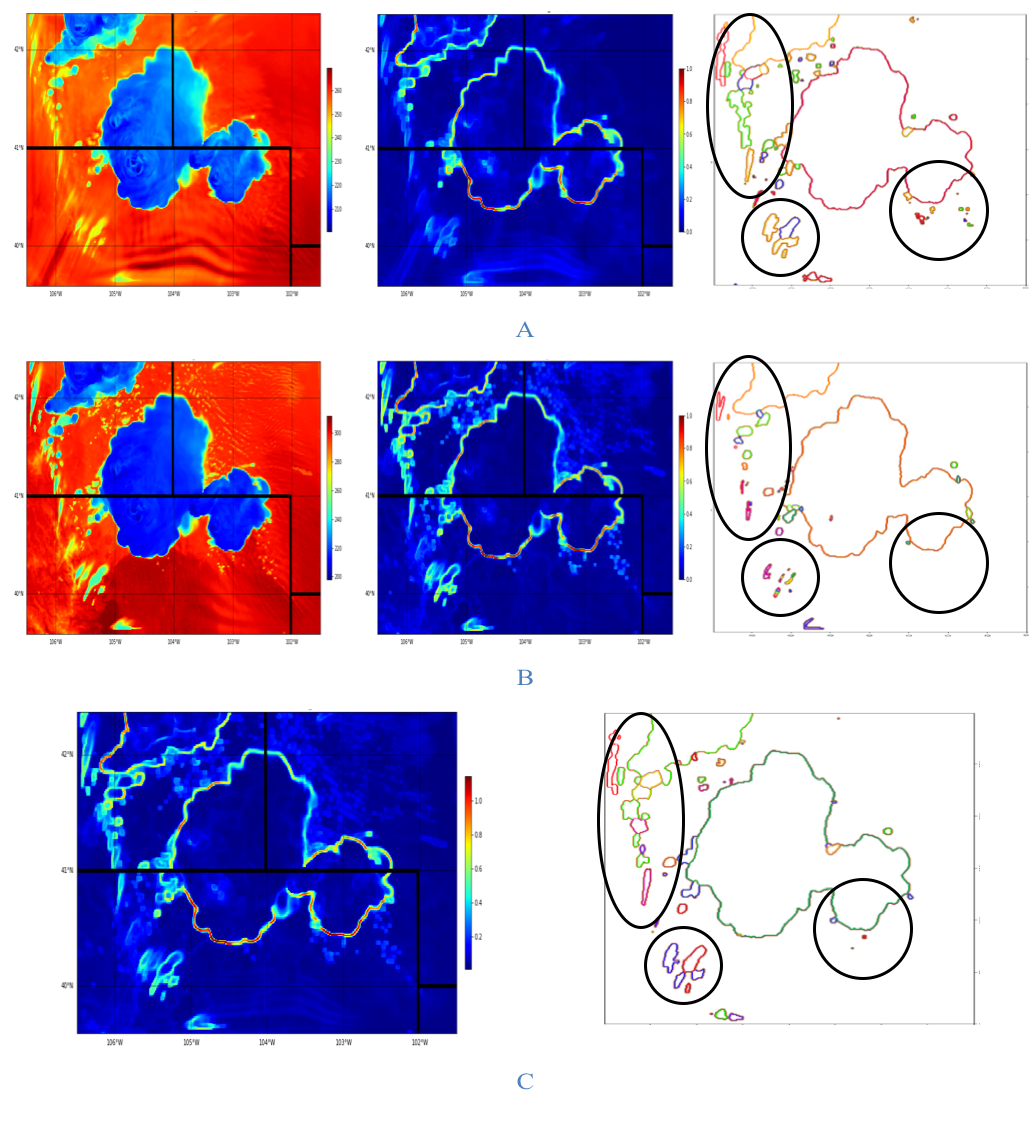}}
 \caption{Visualizing the effect of integrating complementary information from lower-level water vapor for the Wyoming tornado case at 21:05 UTC of the 12th of June 2017: Panels in section A are showing final cloud segments from the gradient magnitudes of lower water vapor band. Panels in section B are showing final cloud segments from the gradient magnitudes of single IR longwave window band. Panels in section C are showing final cloud segments from the combined gradient magnitude image of both spectral bands (IR longwave window and water vapor band).}\label{f10}
\end{figure}

As it is mentioned in the methodology section, the gradient-based segmentation algorithm is capable of taking into account the measurements from different channels of multispectral imagery. In order to visually assess the effect of integrating other complementary channels of information, lower water vapor channel data is also processed and gradient magnitudes are calculated separately for this band (Fig. \ref{f10}). The gradient image values from each spectral bands are then summed up to generate a single-valued multispectral gradient magnitude image. Then the watershed segmentation is applied to the combined single-valued gradient magnitude imagery with the same procedures explained before. The segmentation results from each scenario (IR-only, and IR+ Water vapor) are shown in Fig. \ref{f10}. Figures are implying that the integration of information from other spectral bands will provide more useful information for delineating the clouds distinctly and to discard cloud patches that are mistakenly detected due to the background noise in utilization of only-IR channel data. Adding the water vapor channel provides useful information for better defining the cloud segments and leads to a more accurate classification as the next step toward more robust intensity estimation and areal delineation of rainfall.

\subsection*{Statistical Evaluation}

In order to perform an accuracy assessment, and to compare the outcome of segmentation algorithms, a reference mask is needed. Since there is no accurate "truth" observation mask to determine whether a grid point is correctly covered by cloud or not, the model simulation of mixing ratios of all hydrometeors, including cloud water, cloud ice, rainwater, snow, and graupel, are used. These hydrometeor mixing ratios are prognostic model variables simulated by the Thompson microphysics scheme and evolve in correspondence with the dynamical and thermodynamical processes within the WRF model. After summing them all together and taking the vertical maximum value within each column, a threshold of 10-6 kg/kg is then applied to the horizontal hydrometeor mixing ratio map, which is a threshold widely used for cloud top and cloud base identification from model simulations beginning from \citep{otkin2008comparison}. The grid point is cloudy if hydrometeor mixing ratio at this location is greater than the threshold and is clear sky if lower than the threshold. The model-derived horizontal distribution of clouds is used as the “truth” to verify and compare the two different cloud identification algorithms. The metrics used to assess the cloud segmentation algorithms are based on table \ref{t1}. The verification indices used along with their application are listed in table \ref{t2}.

\begin{figure}[h]
\centerline{\includegraphics[width=29pc]{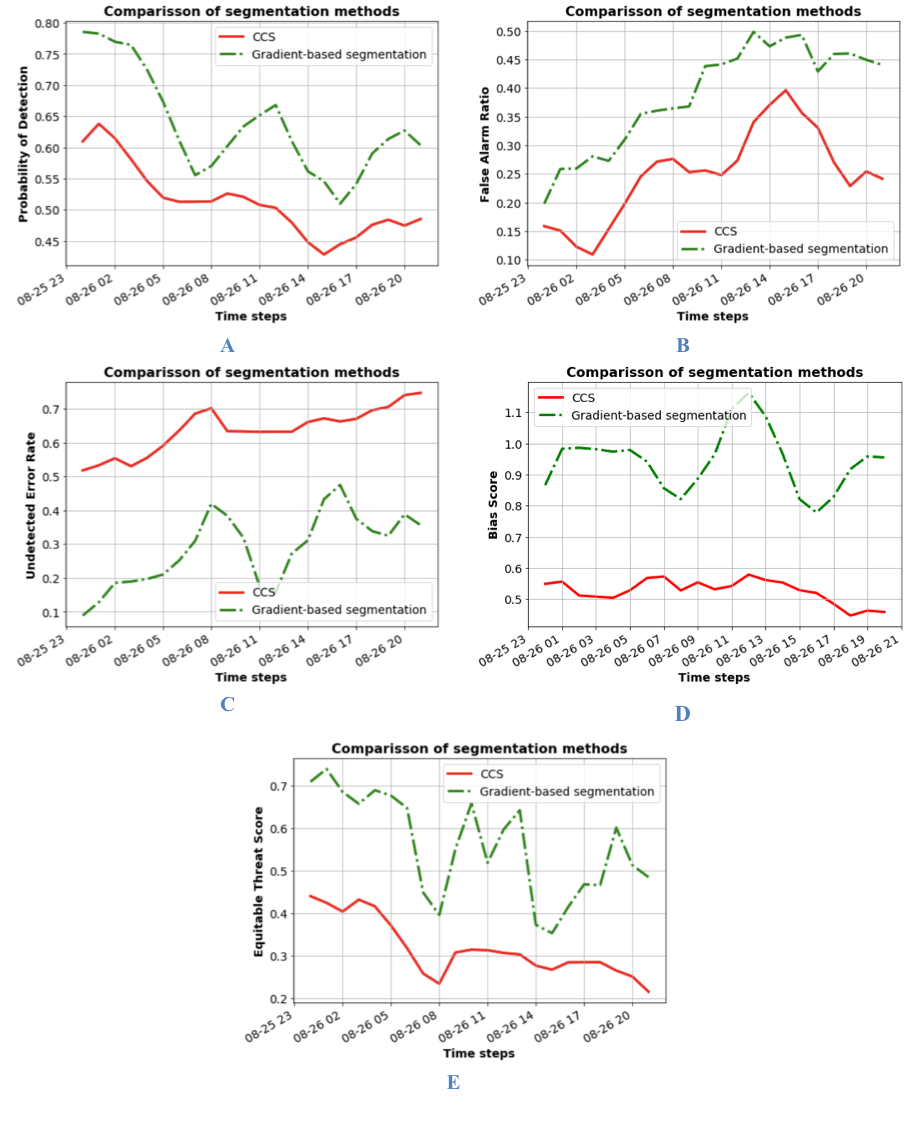}}
 \caption{Statistical comparison of two different segmentation algorithms for Hurricane Harvey case: a. POD, b. False Alarm Ratio, c. Undetected Error Rate, d. Bias Score, e. Equitable Threat Score.}\label{f11}
\end{figure}

\begin{figure}[h]
\centerline{\includegraphics[width=29pc]{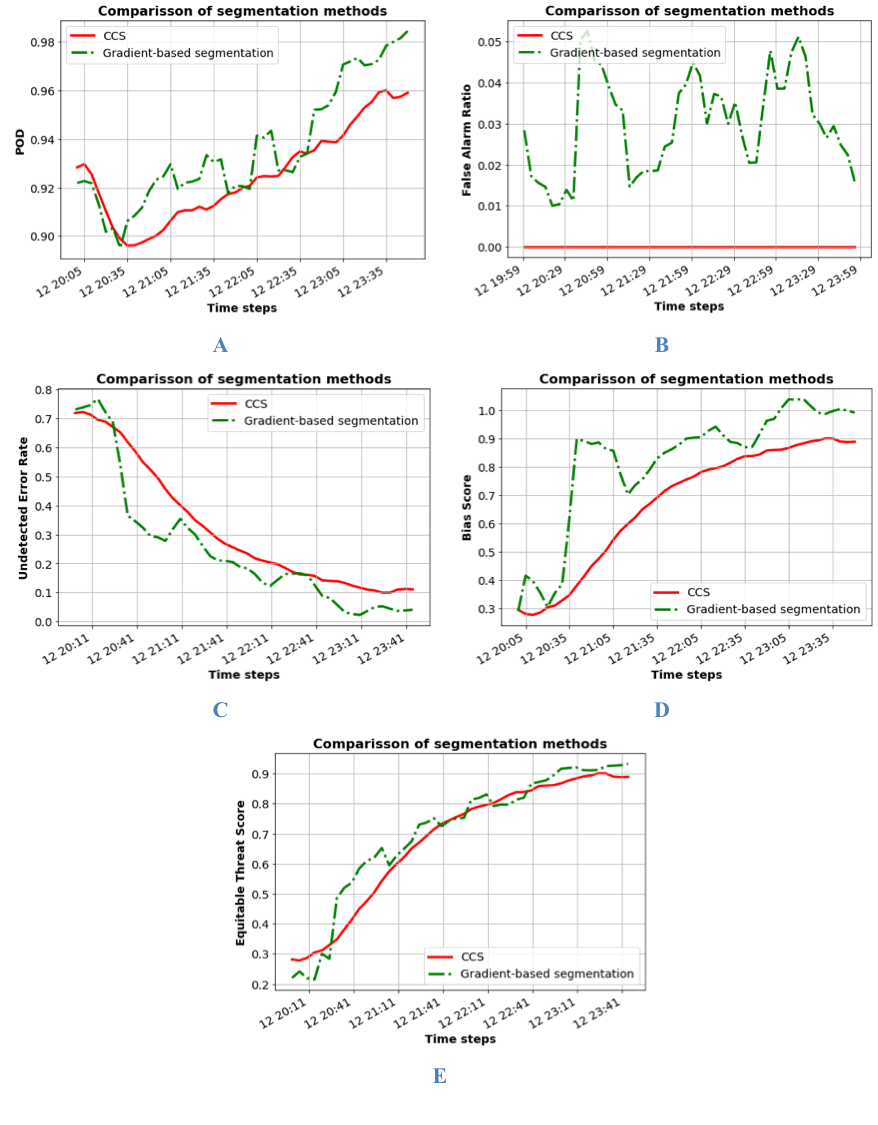}}
 \caption{Statistical comparison of two different segmentation algorithms for Wyoming Tornado case: a. POD, b. False Alarm Ratio, c. Undetected Error Rate, d. Bias Score, e. Equitable Threat Score.}\label{f12}
\end{figure}

Performance improvement in segmentation skill using the gradient-based technique is evident (Fig. \ref{f11}, \ref{f12}). The improvement is pronounced when comparisons are made for using IR-only data to keep the consistency between the proposed segmentation algorithm, and the conventional one implemented in PERSIANN-CCS. The only metric that is not showing the performance enhancement is the False Alarm Ratio (FAR with the best score of zero) which the gradient-based segmentation is showing non-zero but insignificant average value of 0.03.

\section*{Summary and Conclusions}
A gradient-based segmentation algorithm for cloud detection and segmentation using geostationary satellite infrared imageries is presented. The goal is to provide an automated method to overcome the shortcomings associated with the traditional patch-based cloud segmentation approaches toward more reliable precipitation retrievals. This algorithm is based on mathematical morphology and is utilized to extract the information from the single band or more than one channel of satellite imagery. Due to the particular characteristics of each spectral band measurements, accumulation of additional sources of information from multi-channel satellite Imagery is viable by looking at their gradient magnitudes instead of directly utilizing each channel's values. This gradient-based cloud image segmentation method integrates morphological image gradient magnitudes to separable cloud systems and patches boundaries using a convolution operation. The proposed algorithm, as well as the conventional patch-based segmentation approach used in PERSIAN-CCS algorithm, is applied on the simulated GOES-16 ABI imageries accompanied with their modeled horizontal distribution of hydrometeors as the reference to point out the performance enhancement. Experiments and the results from the visual and statistical comparison indicate the improved performance of the gradient-based segmentation technique over the traditional approaches specifically in terms of extracting information useful to identify warm cloud region. This finding is the first step toward reducing the uncertainty associated with the patch-based precipitation retrieval algorithms. 

\begin{table}[h]
\caption{The four possible diagnosis results of testing.}\label{t1}
\begin{center}
\begin{tabular}{|c|c|c|c|}
\hline
                      & \textbf{Detected}                                                                                                                 & \textbf{Not detected}                                                                                                            & \textbf{Probability} \\ \hline
\textbf{Existing}     & \begin{tabular}[c]{@{}c@{}}True positive (TP) \\ The existing defect is detected \\ (hit)\end{tabular}                            & \begin{tabular}[c]{@{}c@{}}False negative (FN)\\ The existing defect is not detected (miss)\end{tabular}                         & TP + FN = 100 \%     \\ \hline
\textbf{Non-existing} & \begin{tabular}[c]{@{}c@{}}False positive (FP)\\ A defect is detected even though it is not existing\\ (False Alarm)\end{tabular} & \begin{tabular}[c]{@{}c@{}}True negative (TN)\\ No defect is detected, where no defect exists\\ (Correct Rejection)\end{tabular} & FP + TN = 100 \%     \\ \hline
\textbf{Total}        & FP + TP                                                                                                                           & FN + TN                                                                                                                          & Total =TP+ FN+TN+FP  \\ \hline
\botline
\end{tabular}
\end{center}
\end{table}

\begin{table}[h]
\caption{Verification Metrics}\label{t2}
\begin{center}
\begin{tabular}{|p{3cm}|p{5cm}|p{2cm}|p{6cm}|}

\hline
\textbf{Verification Metrics}         & \textbf{Formulation}                                                                                    & \textbf{Range}                                                                  & \textbf{Application}                                                                                                       \\ \hline
\textbf{Undetected Error Rate}        & Ur= Misses/Number of observed events                                                                    & \begin{tabular}[c]{@{}c@{}}(0 $\leq$ Ur $\leq$ 1)\\ Perfect score: 0\end{tabular}     &  The rate of error in detection of hit events                                                                                              \\ \hline
\textbf{Probability of Detection}     & POD =Hits/ Number of observed events                                                                    & \begin{tabular}[c]{@{}c@{}}(0 $\leq$ Hr $\leq$ 1)\\ Perfect score: 1\end{tabular}     & \begin{tabular}[c]{@{}c@{}} The likelihood of correct detection\end{tabular}         \\ \hline
\textbf{False Alarm Ratio}             & Fr = False Alarm/Number of Not Observed Events                                                          & \begin{tabular}[c]{@{}c@{}}(0 $\leq$ Fr $\leq$ 1)\\ Perfect score: 0\end{tabular}     & The number of false alarms per total number of alarms                                                                        \\ \hline
\textbf{Bias Score}                   & Bias=(Hits+False alarms)/ Number of observed events                                                     & \begin{tabular}[c]{@{}c@{}}(0 $\leq$ BI)\\ Perfect score: 1\end{tabular}           & \begin{tabular}[c]{@{}c@{}} How similar were the frequencies of Existing\\ and Detected events?\end{tabular}        \\ \hline
\textbf{Equitable Threat Score (ETS)} & \begin{tabular}[c]{@{}c@{}}ETS=Hits-Hitsrandom/(Hits+Misses+False\\ alarms-Hitsrandom)\end{tabular} & \begin{tabular}[c]{@{}c@{}}(-1/3 $<$ ETS $<$ 1)\\ Perfect score: 1\end{tabular} & \begin{tabular}[c]{@{}c@{}} How well did the Existing events\\ correspond to the Detected events\end{tabular} \\ \hline
\botline
\end{tabular}
\end{center}
\end{table}

\acknowledgments
The financial support of this research is from U.S. Department of Energy (DOE Prime Award number DE-IA0000018), California Energy Commission (CEC Award number 300-15-005), MASEEH fellowship, NSF CyberSEES Project (Award CCF-1331915), NOAA/NESDIS/NCDC (Prime Award NA09NES4400006 and subaward 2009-1380-01), the NASA Earth and Space Science Fellowship (Grant NNX15AN86H). The participation of the Pennsylvania State University co-authors is supported by NASA Grants NNX16AD84G and NNX12AJ79G. The Harvey WRF forecast used here is conducted and provided by Masashi Minamide. Data assimilation and numerical simulations were performed on the Stampede supercomputer of the Texas Advanced Computing Center (TACC). Finally, authors would like to sincerely thank the valuable comments and suggestion of the editors and the anonymous reviewers.


 \bibliographystyle{ametsoc2014}
 \bibliography{neginHayatbini}






\end{document}